\documentclass[sigconf,natbib=true,balance=false]{acmart}

\AtBeginDocument{%
  \providecommand\BibTeX{{%
    \normalfont B\kern-0.5em{\scshape i\kern-0.25em b}\kern-0.8em\TeX}}}

\copyrightyear{2023}
\acmYear{2023}
\setcopyright{acmlicensed}\acmConference[SIGIR '23]{Proceedings of the 46th International ACM SIGIR Conference on Research and Development in Information Retrieval}{July 23--27, 2023}{Taipei, Taiwan}
\acmBooktitle{Proceedings of the 46th International ACM SIGIR Conference on Research and Development in Information Retrieval (SIGIR '23), July 23--27, 2023, Taipei, Taiwan}
\acmPrice{15.00}
\acmDOI{10.1145/3539618.3591763}
\acmISBN{978-1-4503-9408-6/23/07}

\usepackage[ruled,lined,boxed,commentsnumbered]{algorithm2e}
\usepackage{microtype}
\usepackage{multirow}
\usepackage{multicol}
\usepackage{amsmath}
\usepackage{booktabs,subcaption,amsfonts,dcolumn}
\usepackage{color} 
\usepackage{url}
\usepackage[utf8]{inputenc}
\usepackage{graphicx}

\newcommand\blank{$\rule{0.5cm}{0.15mm}$}
\newcommand\nw[1]{``\texttt{\small #1}''}
\SetKwRepeat{Do}{do}{while}%
\newcommand{\ours}{RAP~}
\acmSubmissionID{7131}
\usepackage[most]{tcolorbox}

\begin{document}

\title{Schema-aware Reference as Prompt Improves Data-Efficient Knowledge Graph Construction}

%


%

\author{Yunzhi Yao}
\authornotemark[1]
\affiliation{%
  \institution{Zhejiang University \\ AZFT Joint Lab for Knowledge Engine }
  \state{Zhejiang}
  \country{China}
}
\email{yyztodd@zju.edu.cn}

\author{Shengyu Mao}
\authornote{Equal contribution and shared co-first authorship.}
\affiliation{%
  \institution{Zhejiang University \\ AZFT Joint Lab for Knowledge Engine }
  \state{Zhejiang}
  \country{China}
}
\email{shengyu@zju.edu.cn}

\author{Ningyu Zhang}
\authornotemark[2]
\affiliation{%
  \institution{Zhejiang University \\ AZFT Joint Lab for Knowledge Engine}
  \state{Zhejiang}
  \country{China}
}
\email{zhangningyu@zju.edu.cn}

\author{Xiang Chen}
\affiliation{%
  \institution{Zhejiang University \\ AZFT Joint Lab for Knowledge Engine}
  \state{Zhejiang}
  \country{China}
}
\email{xiang_chen@zju.edu.cn}

\author{Shumin Deng}
\affiliation{%
  \institution{National University of Singapore \\ NUS-NCS Joint Lab}
  \country{Singapore} 
}
\email{shumin@nus.edu.sg}

\author{Xi Chen}
\affiliation{%
  \institution{Tencent}
  \state{Guangdong}
  \country{China}
}
\email{jasonxchen@tencent.com}

\author{Huajun Chen}
\authornote{Corresponding author.}
\affiliation{%
  \institution{Zhejiang University \\ AZFT Joint Lab for Knowledge Engine }
   \institution{Donghai Laboratory }
  \state{Zhejiang}
  \country{China}
}
\email{huajunsir@zju.edu.cn}




\renewcommand{\shortauthors}{Yunzhi Yao, et al.}

\begin{abstract}
With the development of pre-trained language models, many prompt-based approaches to data-efficient knowledge graph construction have achieved impressive performance. However, existing prompt-based learning methods for knowledge graph construction are still susceptible to several potential limitations: (i) semantic gap between natural language and output structured knowledge with pre-defined schema, which means model cannot fully exploit semantic knowledge with the constrained templates; (ii) representation learning with locally individual instances limits the performance given the insufficient features, which are unable to unleash the potential analogical capability of pre-trained language models. Motivated by these observations, we propose a retrieval-augmented approach, which retrieves schema-aware \textbf{R}eference \textbf{A}s \textbf{P}rompt (\textbf{RAP}), for data-efficient knowledge graph construction. It can dynamically leverage schema and knowledge inherited from human-annotated and weak-supervised data as a prompt for each sample, which is model-agnostic and can be plugged into widespread existing approaches. Experimental results demonstrate that previous methods integrated with \textsc{RAP} can achieve impressive performance gains in low-resource settings on five datasets of relational triple extraction and event extraction for knowledge graph construction\footnote{Code is available in \url{https://github.com/zjunlp/RAP}.}.
\end{abstract}


\maketitle

\section{Introduction}
Knowledge Graphs (KGs) as a form of structured knowledge can provide back-end support for various practical applications, including information retrieval \cite{wise2020covid}, question answering \cite{DBLP:journals/corr/abs-2007-13069}, and recommender systems \cite{wang2019kgat,cao2019unifying}. 
Knowledge graph construction aims to automatically retrieve specific relational triples and events from texts \cite{DBLP:journals/tnn/JiPCMY22}.
Most prior works on knowledge graph extraction rely on a large amount of labeled data for training \cite{DBLP:conf/acl/ZhengWCYZZZQMZ20}; however, high-quality annotations are expensive to obtain.
Thus, many data-efficient approaches have been proposed \cite{DBLP:conf/www/ChenZXDYTHSC22}, in which prompt-based learning with Pre-trained Language Models (PLMs) yields promising performance.
For example, \cite{DBLP:conf/acl/ChiaBPS22} designs a structured prompt template for generating synthetic relation samples for data-efficient relational triple extraction. 
\cite{DBLP:conf/naacl/HsuHBMNCP22} formulates event extraction as a conditional generation problem with a manually designed prompt, which achieves high performance with only a few training data.

\begin{figure}
    \includegraphics[width=0.4\textwidth]{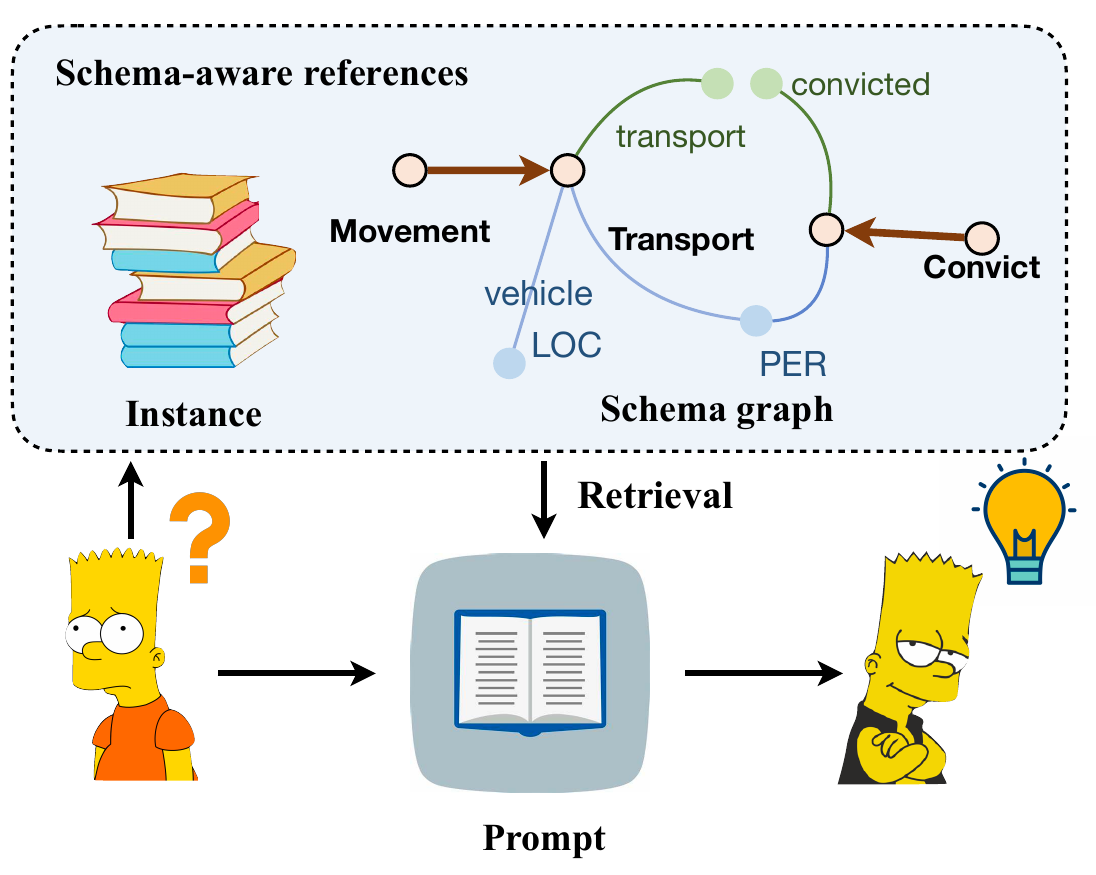}
    \caption{
    Schema-aware reference as prompt. 
    We construct a schema-instance hybrid reference store from which 
    we retrieve related knowledge as a prompt for data-efficient learning with PLMs (e.g., BART \cite{lewis2020bart}).}
    \label{fig:overview}
\vspace{-2mm}
\end{figure}

Existing methods have notable limitations. Unlike general NLP tasks, knowledge graph construction requires structured prediction that adheres to a pre-defined schema.
Raw text data for PLMs may not have sufficient task-specific patterns, leading to a semantic gap between the input sequence and schema.
Constrained prompt templates struggle to fully utilize semantic knowledge and generate schema-conforming outputs.
Moreover, prior prompt-based learning relies on the parametric-based paradigm, which is unable to unleash the potential analogical
capability of pre-trained language models \cite{DBLP:conf/nips/BrownMRSKDNSSAA20}.
Notably, they may fail to generalize well for complex examples and perform unstably with limited training data since the scarce or complex examples are not easy to be learned in parametric space during optimization.
For example, texts mentioning the same event type can vary significantly in structure and expression. 
``\textit{A man was hacked to death by the criminal}'' and ``\textit{The aircraft received fire from an enemy machine gun}'' both describe an \emph{Attack} event, although they are almost literally different.
With only few-shot training samples, the model may struggle to discriminate such complex patterns and extract correct information.

To overcome the aforementioned limitations, we try to fully leverage the schema and global information in training data \textbf{as references} for help.
Note that humans can use associative learning to recall relevant skills in memories to conquer complex tasks with little practice.
Similarly, given the insufficient features of a single sentence in the low-resource setting, it is beneficial to leverage that schema knowledge and the similar annotated examples to enrich the semantics of individual instances and provide reference \cite{DBLP:conf/naacl/WangCZCLH22}.
Motivated by this, as shown in Figure \ref{fig:overview}, we propose a novel approach of schema-aware \textbf{R}eference \textbf{A}s \textbf{P}rompt (\textbf{RAP}), which dynamically leverages symbolic schema and knowledge inherited from examples as prompts to enhance the PLMs for knowledge graph construction. 

However, there exist two problems: 
(1) \emph{Collecting reference knowledge:} 
Since rich schema and training instances are complementary to each other, it is necessary to combine and map these data accordingly to construct reference store.
(2) \emph{Leveraging reference knowledge:} Plugin-in-play integrating those reference knowledge to existing KG construction models is also challenging since there are various types of models (e.g., generation-based and classification-based methods).

To address the problem of collecting reference knowledge, we propose a \emph{schema-aware reference store} that enriches schema with text instances. 
Specifically, we align instances from human-annotated and weak-supervised text with structured schema; thus, symbolic knowledge and textual corpora are in the same space for representation learning.
Then we construct a unified reference store containing the knowledge derived from both symbolic schema and training instances. 
To address the problem of leveraging reference knowledge, we propose \emph{retrieval-based reference integration} to select informative knowledge as prompts \cite{DBLP:conf/www/YeZDCCXCC22}. 
Since not all external knowledge is advantageous, we utilize a retrieval-based method to dynamically select knowledge as prompts that are the most relevant to the input sequence from the schema-aware reference store.
In this way, each sample can achieve diverse and suitable knowledgeable prompts that can provide rich symbolic guidance in low-resource settings.

To demonstrate the effectiveness of our proposed \textbf{RAP}, we apply it to knowledge graph construction tasks of relational triple extraction and event extraction tasks.
Note that our approach is model-agnostic and readily pluggable into any previous approaches.
We evaluate the model on two relation triple extraction datasets: NYT and WebNLG, and two event extraction datasets: ACE05-E and CASIE.
Experimental results show that the \textsc{RAP} model can perform better in low-resource settings. 



\begin{figure*}       
    \centering
    \includegraphics[width=0.91 \textwidth]{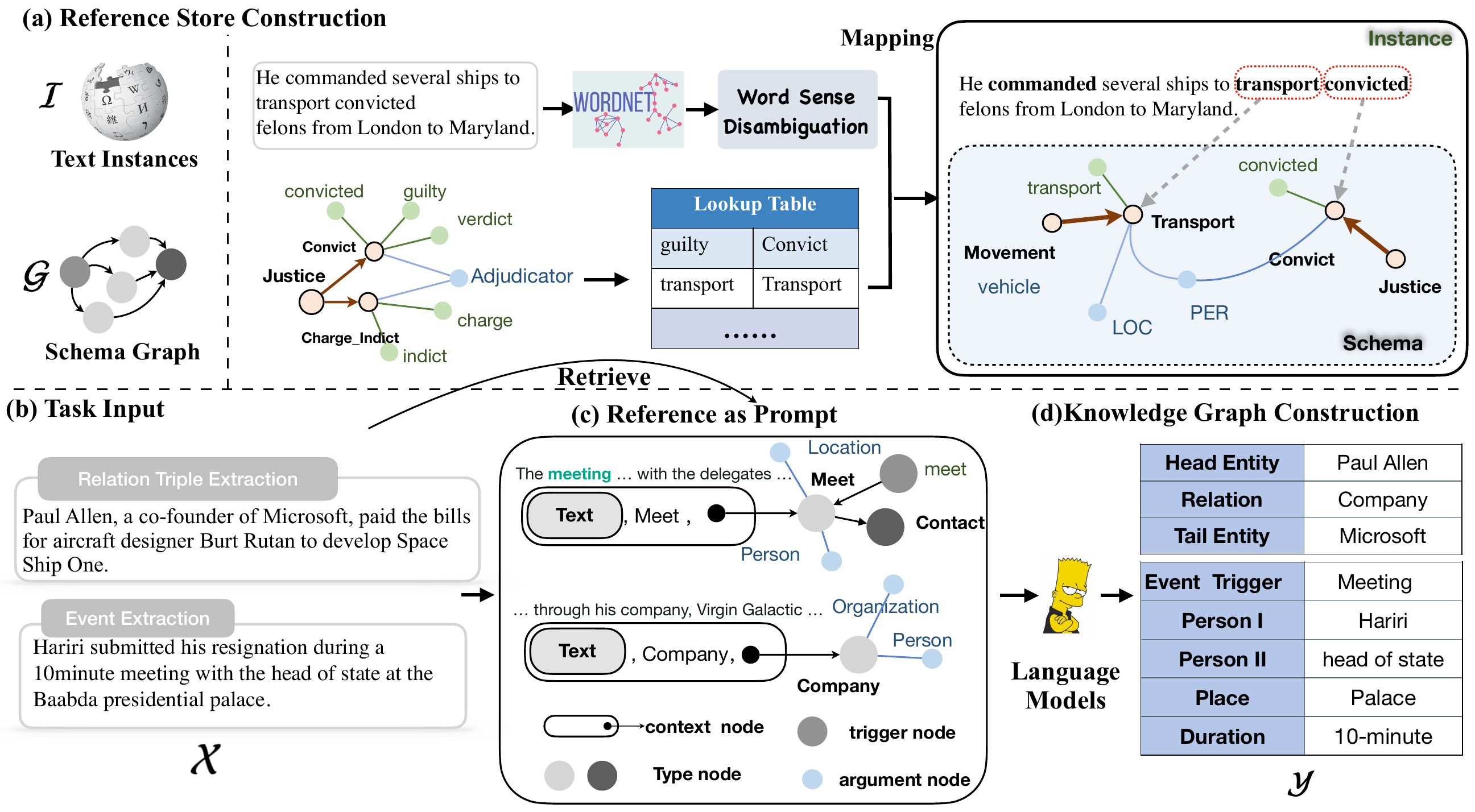}
    \caption{The architecture of schema-aware \textbf{R}eference \textbf{A}s \textbf{P}rompt (\textsc{RAP}), which is model-agnostic and is readily pluggable into many existing KGC approaches \textsc{TEXT2EVENT} \cite{DBLP:conf/acl/0001LXHTL0LC20}, \textsc{DEGREE} \cite{DBLP:conf/naacl/HsuHBMNCP22}, \textsc{PRGC} \cite{DBLP:conf/acl/ZhengWCYZZZQMZ20}, \textsc{RelationPrompt} \cite{DBLP:conf/acl/ChiaBPS22} and so on.}
    \label{fig:pipeline}
\end{figure*}

\section{Preliminaries}
In this paper, we apply our approach, \textbf{RAP}, to two representative tasks of knowledge graph construction, namely: relation triple extraction and event extraction. 
\subsection{Task Definition}
\textbf{Event Extraction.} Event extraction is the process of automatically extracting events from unstructured natural language texts, guided by an event schema. 
To clarify the process, the following terms are used: a trigger word is a word or phrase that most accurately describes the event, and an event argument is an entity or attribute involved in the event, such as the time or tool used.
For example, the sentence ``A man was hacked to death by the criminal'' describes an \emph{Attack} event triggered by the word `hacked'. 
This event includes two argument roles: the \emph{Attacker} (criminal) and the \emph{Victim} (a man). 
The model should be able to identify event triggers, their types, arguments, and their corresponding roles.

\textbf{Relation Triple Extraction.} Joint extraction of entity mentions and their relations which are in the form of a triple (subject, relation, object) from unstructured texts, is an important task in knowledge graph construction. 
Given the input sentences, the desired outputs are relational triples $(e_{head}, r, e_{tail})$, where $e_{head}$ is the head entity, $r$ is the relation, and $e_{tail}$ is the tail entity.
For instance, given the sentence ``His 35-year career at Moil Oil included a four-year assignment in Tokyo, Japan as head of Mobil Far East.'', the model should identify two entities \emph{Tokyo} and \emph{Japan} and their relation \emph{capital-of}, described as triple \emph{(Tokyo, capital-of, Japan)}. 

\subsection{Problem Formulation}
Given an original text $X$, the purpose of the information extraction task is to obtain target information $\mathcal{Y}=\{\mathcal{Y}^1, ..., \mathcal{Y}^t \}$, where $\mathcal{Y}^i, i \in t$ represents the information to extract for the j-th type, and $t$ refer to the number of types.
For the relation triple extraction task, $\mathcal{Y}^i$ is in the form of triples $\mathcal{Y}^i=(e_{head}, r, e_{tail})$, including the head entity, tail entity, and their relation. 
For the event extraction, $\mathcal{Y}^i$ contains the corresponding event record in the sentence, which can be represented as $\mathcal{Y}^i = \{event-type, trigger, argument-role\}$.
In the following part, we will introduce the prompt construction and application details.


\section{Methodology}
Figure~\ref{fig:pipeline} illustrates the framework of \textsc{RAP}.
We collect knowledge from different sources and construct a schema-aware reference store (Section {\ref{sec:store}}). 
Then, we dynamically retrieve related references for each query as the prompt to inject into the model~(Section {\ref{sec:injection}}).

\subsection{Schema-aware Reference Store Construction} 
\label{sec:store}
 
\subsubsection{\textbf{Base Reference Store}} 
The base reference store contains the text instances $\mathcal{I}$ which contain a wealth of information that may share semantic similarities with the query $\mathcal{X}$.
A well-sized retrieval source is crucial for the text instances, as too large of a textbase can lead to noise and increased search space, while too small of a textbase would be ineffective.
Previous research \cite{DBLP:conf/acl/WangXFLSX0022} indicates that using training data as the datastore can improve downstream tasks; therefore, we use training data to construct the base reference store.

\subsubsection{\textbf{Schema-instance Hybrid Reference Store}}
Since the base reference store does not contain any structure schema knowledge; we employ schema information to augment the references.
A task schema is a symbolic graph $\mathcal{G}$ describing the configuration of each target type.
As demonstrated in Figure~\ref{fig:pipeline}, these nodes (knowledge types) are connected through their intrinsic relationships. 
Taking the event extraction task as an example, the event `\emph{meet}' is linked with `\emph{Meet}' since `meet' is a trigger word for the \emph{Meet} event. 
For the event extraction task, the schema graph includes three types of nodes: the event type $\mathcal{E}$, trigger word $\mathcal{T}$, and argument role $\mathcal{A}$.
We follow previous work~\cite{DBLP:conf/acl/LinJHW20, DBLP:conf/emnlp/LiZLCJMCV20,DBLP:conf/acl/DaganJVHCR18}  and leverage the event schema\footnote{\url{www.ldc.upenn.edu/sites/www.ldc.upenn.edu/files/english-events-guidelines-v5.4.3.pdf}} provided by the dataset. 
For the relational triple extraction task, the schema graph contains both the relation type $\mathcal{R}$ and the entity information $\mathcal{S}$, and we build the schema graph based on the original dataset such as WebNLG or NYT.
The base reference store contains the labeled training data and we link the text instance to the schema graph $\mathcal{G}$ based on the label.

Note that the size of the schema-aware reference store is based on the number of annotated training data; however, high-quality data is usually scarce due to the expensive cost of annotation in low-resource scenarios.
Since previous work \cite{min2022rethinking} has demonstrated that randomly replacing labels in the demonstrations \emph{barely hurts the classification performance} while the key lies in  the label space and the distribution
of the input text.
Inspire by this, we take the first step to extend the reference store with weak-supervised open domain corpus, which will be introduced in the following sections.

\subsubsection{\textbf{Reference Store Extension with Weak Supervision}}
We introduce the reference store extension method using event extraction as an example, which is readily simple to apply to any other knowledge graph construction tasks.
We use Wikipedia as the external data for extension and select the subset of Wikipedia EventWiki~\cite{ge-etal-2018-eventwiki} and another event data \cite{DBLP:conf/acl/ChenLZLZ17} from FreeBase.
Concretely, we automatically ``annotate'' weak labels for these corpora and add them into the reference store.

Here, we introduce a simple knowledge-guided weak supervision method to tag potential events in the sentence.
Given a sentence $c$ from the corpora $C$, we aim at inducing its potential label $\tilde{y}\in \mathcal{Y}$ and link $s$ to the graph $\mathcal{G}$ based on $\tilde{y}$.
Here, we adopt a lightweight symbolic pipeline method named Trigger From WordNet (TFW) as proposed by~\cite{DBLP:conf/acl/TongXWCHLX20} to ``annotate'' candidate triggers in each sentence. 
Particularly, we first apply IMS~\cite{zhong-ng-2010-makes} to disambiguate words into word sense in WordNet~\cite{10.1093/ijl/3.4.235}. 
Then, we implement the simple dictionary-lookup approach from ~\citet{Araki2018OpenDomainED} to determine whether the word sense triggers an event. 
After we obtain these candidate triggers, we traverse these triggers to find out whether they can be mapped to the target schema graph $\mathcal{G}$. 
We list the detailed steps in Algorithm~\ref{algo:wl}.
\begin{algorithm}  
  \caption{Knowledge-guided Weak Supervision}  
  \label{algo:wl}
  \KwIn{Unlabeled Corpora $C=\{c_1,c_2,..,c_i\}$, WordNet $wn$, Schema Graph $\mathcal{G}$} 
  \KwOut{$\tilde{y}$ for each $c_i \in C$}
  \BlankLine
  \For{$c_i \in C $}{
    $\{(tk_i,pos_i,sense_i)\} \leftarrow $  IMS($c_i, wn$)\;
    $tk_i$ is a token, $pos_i$ is the pos tag\;
    filter $tk_i$ whose $sense_i$ is None \;
    \For{$(tk_i,pos_i,sense_i)$}{
        \If{$pos_i$ \textbf{in} ['verb','noun']}{
            \If{Lookup($sense_i$) is True}{
                $nugget_i \leftarrow$ CreateEvent($tk_i$)
            }
        }
    }
    \For{$nugget_i$}{
        $\tilde{y}_{i} \leftarrow $ Mapping($c_i, nugget_i, T$)
    }
  }
\end{algorithm}

For the example illustrated in Figure~\ref{fig:pipeline}, given the sentence $k$ = \emph{``He commanded several ships contracted by Jonathan Forward to transport convicted felons from London to Maryland.''}, we can obtain the candidate triggers \emph{("commanded", "contracted", "transport", "convicted")}. 
We then map the trigger ``transport'' to $\tilde{y}_1= $ \emph{``Transport''} and the ``convicted'' to $\tilde{y}_2=$ \emph{``Convict''} in the schema graph. 
It can be noted that the sentence does not contain a \emph{``Convict''} event since the word ``convicted'' is just an attribute used to decorate the following word ``felons''. 
However, here we just require a weak label to extend the schema-aware reference store; thus, those errors are tolerable. 
Notably, the simple knowledge-guided weak supervision can be flexibly applied to other knowledge graph construction tasks with unsupervised corpus.

Finally, we store the schema-aware reference store as a key-value memory:  
\begin{itemize}
  \item \textbf{keys}: the entries (key) of the knowledge store are the text instances;
  \item \textbf{values}: the pointers that can be linked to one or several nodes in the schema graph.
\end{itemize}

Specifically, as shown in Figure~\ref{fig:pipeline}, given the i-th instance $c_i$, every entry is stored as $(c_i, \tilde{y}_i, p_i)$, where $c_i$ is the context, $\tilde{y}_i$ is the label and $p_i$ is a pointer that is linked to the type nodes in the schema graph. 
Additionally, we leverage triples to store the schema graph, such as $(\text{Meet}, SubType, \text{Contact})$.
\begin{equation}
\mathcal{D}=\{(c_i,\tilde{y}_i,p_i) \}
\end{equation}

\subsection{Retrieval-based Reference Integration}
\label{sec:injection}

\ours~ construct a unique and suitable prompt for each sample by retrieving knowledge from the schema-aware reference store and integrating it with the input to feed into the knowledge graph construction model.
Formally, we have:
\begin{equation}
p_{\text {}}(\mathcal{Y} \mid \mathcal{X}) \approx \sum_{\mathcal{Z} \in \text { top }-k(p(\cdot \mid \mathcal{X}))} p_\eta(\mathcal{Z} \mid \mathcal{X}) p_\theta(\mathcal{Y} \mid \mathcal{X}, \mathcal{Z})
\end{equation}
where $\mathcal{Z}$ refers to the reference as prompt retrieved from the store.
The retrieval component $p_\eta(\mathcal{Z} \mid \mathcal{X})$ is based on an off-the-shelf sparse searcher based on Apache Lucene, Elasticsearch, using an inverted index lookup. 
Specially, we query the reference store with the input text $x$, the engine computes the scores with each item in the store according to a similarity function $score(.,.)$ as follows:
\begin{equation}
    score(x,d_{i})=|x, d_{i}|_{2},
\end{equation}
where $d_{i} \in \mathcal{D}$, and $||_{2}$ indicates to BM25 score.
In this way, we can obtain the top $k$ most similar entries $\{{d}_1,..,d_k\}, d_i=(c_i, \tilde{y}_i, p_i)$ from the datastore. 
We collect the instances $c$ and the schema sub-graphs connected to the pointers $p_i$.

For even extraction, we construct the prompt $\mathcal{Z}_e$ based on the following parts:
(1) Event type $\mathcal{E}$: event type hypernym relation and its definition.
(2) Trigger Information $\mathcal{T}$: we randomly select three trigger nodes that are connected to the event-type node and formulate the trigger prompt as ``Similar trigger such as ...''.
(3) Argument Information $\mathcal{A}$: we follow \citet{DBLP:conf/naacl/HsuHBMNCP22} to build the argument descriptions based on the argument nodes. 
(4) Text Instance $\mathcal{I}$: the final part of the prompt is the text instances we retrieved.
We combine the different knowledge together as the prompt $\mathcal{Z}_e$ as follows:
\begin{equation}
   \mathcal{Z}_e  = Concat(\mathcal{E},\mathcal{T}, \mathcal{A}, \mathcal{I}) 
\end{equation}
For relational triple extraction, we construct the prompt $\mathcal{Z}_r$  similar to $\mathcal{Z}_e$ but contains the following parts:
(1) Relation type $\mathcal{R}$: relation type demonstrating the potential relation that may be described in the sentence.
(2) Structure Information $\mathcal{S}$: the structure information indicates the entity type that formulates the triple such as (city, capital\_of, city).
(3) Text Instance $\mathcal{I}$: the final part of the prompt is the text instances we retrieved above.
We combine the different knowledge together as the prompt $\mathcal{Z}_r$ as follows:
\begin{equation}
    \mathcal{Z}_r  = Concat(\mathcal{R},\mathcal{S}, \mathcal{I}) 
\end{equation}

\subsection{Training and Inference}
\label{train_infer}

After obtaining the prompts for each sample, we apply them in different ways, including both generation-based and classification-based models. 
Previous baselines usually leverage classification-based architectures \cite{DBLP:conf/acl/ZhengWCYZZZQMZ20} or formulate the knowledge graph construction task as a conditional generation problem \cite{DBLP:conf/naacl/HsuHBMNCP22,DBLP:conf/acl/ChiaBPS22}.
In this paper, we mainly focus on end-to-end methods since pipelined methods usually demand fine-grained annotations.
We concatenate $\mathcal{Z}$ with the query sentence $\mathcal{X}$ as the model's input.
\begin{equation}
    \textsc{Input}  = [\mathcal{X} ; [SEP]; \mathcal{Z}]
\end{equation}
$[;]$ denotes the sequence concatenation operation and [SEP] is the corresponding separate marker in the applied PLM.

\textbf{Generation-based}: We optimize the model as a conditional generation~\cite{DBLP:conf/naacl/HsuHBMNCP22}.
Suppose $\theta$ denotes the model's training parameters, the training target is to minimize the negative log-likelihood of all target outputs $\mathcal{Y}_{{j}}$ in training set $\mathcal{N}$. 
Formally, we have:
\begin{equation}
    \mathcal{L}_{\theta}(\mathcal{N}) = - \sum_{j=1}^{|\mathcal{N}|} \log p_\theta(\mathcal{Y}_{j}|\textsc{Input}) 
\end{equation}
$\mathcal{Y}_{{j}}$ has different formation according to the applied model.

\textbf{Classification-based}:
Classification model \cite{DBLP:conf/acl/ZhengWCYZZZQMZ20} usually adopts an encoder to obtain the hidden states $\textbf{h}$ of the input and feeds $\textbf{h}$ into the $Classifier$ to detect the label of each token.
Here, we follow the same input format as the generation-based methods.
However, during prediction, the model should ignore the prompt text, which may bias the semantics for extracting specific entities, relations, and events.
To achieve this, we create a $prompt\_mask$ to exclude the prompt text and rely on the original sentence's hidden states $\textbf{h}_{sen}$ for predictions as:
\begin{gather}
  \textbf{h} = Encoder(\textsc{Input}) \\
  \textbf{h}_{sen} = prompt\_mask(\textbf{h}) \\
  p = Classifier( \textbf{h}_{sen})
\end{gather}
The training target in classification model is to minimize the cross-entropy loss of all target outputs $\mathcal{Y}_{i}$ in training set $\mathcal{N}$, $\mathbb{I}(y=\mathcal{Y}_{i})$ is 1 if the prediction is right and 0 otherwise:
\begin{equation}
  \mathcal{L}_{\theta}(\mathcal{N})=-\frac{1}{|\mathcal{N}|} \sum_{i=1}^{|\mathcal{N}|}(\mathbb{I}(y=\mathcal{Y}_{i}) \log p)
\end{equation}
\subsection{Model Analysis}
\label{sec:model-analysis}

\subsubsection{\textbf{Theoretical Discussion}}
\label{sec:theory}

Note that the proposed approach belongs to the family of retrieval-based methods that typically learn a scorer to map an input instance and relevant (labeled) instance to a score vector, including various successful models such as REINA~\citep{wang2022training}, KATE~\citep{liu2022makes}, and RetroPrompt \cite{DBLP:journals/corr/abs-2205-14704}.
Here, we provide an informal theoretical discussion for better understanding.
The target objective is to learn a function $f : \mathcal{X} \times (\mathcal{X} \times \mathcal{Y})^{\star} \to \mathcal{R}^{|\mathcal{Y}|}$.
We can restrict the work to a sub-family of such retrieval-based approaches that first map $\mathcal{R}^\mathcal{X} \sim \mathcal{D}^{\mathcal{X},\mathcal{Z}}$ to $\hat{\mathcal{D}}^{\mathcal{X},\mathcal{Z}}$ --- an empirical estimate of the local distribution $\mathcal{R}^{\mathcal{X},\mathcal{Z}}$, which is subsequently utilized to make a prediction for $\mathcal{X}$. 
Formally, we have:
\begin{small}
\begin{align}
(\mathcal{X}, \mathcal{Z}^{\mathcal{X}}) \mapsto f(\mathcal{X}, \hat{\mathcal{D}}^{\mathcal{X},\mathcal{Z}}) = \big( f_1(\mathcal{X}, \hat{\mathcal{D}}^{\mathcal{X},\mathcal{Z}}),\ldots,f_{|\mathcal{Y}|}(\mathcal{X}, \hat{\mathcal{D}}^{\mathcal{X},\mathcal{Z}})\big) \in \mathcal{Z}^{|\mathcal{Y}|}
\end{align}
\end{small}
Notably, such a retrieval-based mechanism can extend the feature space $\widetilde{\mathcal{X}} := \mathcal{X} \times \Delta_{\mathcal{X}\times \mathcal{Y}}$, providing a very rich class of functions for knowledge graph construction.
Here,  we focus on a specific form of  $\widetilde{\mathcal{X}}$ with similar semantic information and structure inductive bias, adapting this work for better data-efficient knowledge graph construction.
Previous study \cite{DBLP:journals/corr/abs-2210-02617} theoretically demonstrate that the size of the retrieved set $\mathcal{Z}^\mathcal{X}$  has to scale at least logarithmically in the size of the training set $\mathcal{N}$ to ensure convergence. 

\subsubsection{\textbf{Model Computation Cost Analysis}}
\label{sec:time}
Construction of the schema-aware reference store requires a single forward pass
over the training set and external weak supervised data, which amounts to a fraction of memory and model computation.
Once the reference store is constructed, for the ACE05-E dataset building the cache with 8.5M entries takes roughly 10 seconds on a single CPU. 
Finally, the evaluation on the test set takes about 14ms  per instance when retrieving references as prompt.
Note that the computation cost of building a large schema-aware reference store grows linearly with the number of keys, thus it is trivial to parallelize.

\begin{table*}[ht]
\centering
\setlength{\tabcolsep}{3pt}
\caption{
F1-score results for low-resource event extraction (trigger classification and argument classification) on the ACE-05 dataset.
The highest scores are in \textbf{bold}.
Note that results of baseline models are directly taken from DEGREE~\cite{DBLP:conf/naacl/HsuHBMNCP22}.}
\begin{tabular}{l|c|cccccc|cccccc}
    \toprule
    \multirow{2}{*}{\textbf{Model}} & \multirow{2}{*}{\textbf{Type}} & \multicolumn{6}{c|}{\textbf{Tri-C}} & \multicolumn{6}{c}{\textbf{Arg-C}}  \\
    \cline{3-14}
    & & 1\% & 3\% & 5\% & 10\% & 20\% & 30\% & 1\% & 3\% & 5\% & 10\% & 20\% & 30\%  \\
    \midrule
    BERT\_QA & Classification
    & 20.5 & 40.2 & 42.5 & 50.1 & 61.5 & 61.3 
    & 4.7 & 14.5 &  26.9 & 27.6 & 36.7 & 38.8 \\
    \textsc{OneIE} & Classification
    & 38.5 & 52.4 & 59.3 & 61.5 & 67.6 & 67.4
    &  9.4 & 22.0 & 26.8 & 26.8 & 42.7 & 47.8 \\
    \textsc{TANL} & Generation 
    & 34.1 & 48.1 & 53.4 & 54.8 & 61.8 & 61.6
    & 8.5 & 17.2 & 24.7 & 29.0 & 34.0 & 39.2 \\
    \textsc{DEGREE(Pipe)} & Generation
    & 55.1 & 62.8 & 63.8 & 66.1 & 64.4 & 64.4
    &  13.1 & 26.1 & 27.6 & 42.1 & 40.7 & 44.0 \\
    \midrule
    \textsc{Text2Event} & Generation
    & 14.2 & 35.2 & 46.4 & 47.0 & 55.6 & 60.7
    &  3.9 & 12.2 & 19.1 & 24.9 & 32.3 & 39.2 \\
    \textbf{RAP$_{(TEXT2EVENT)}$ } & Generation
    & 19.3  & 36.5  & 47.8 & 48.0  & 59.3  & 62.7 
    & 7.2  & 16.2  & 19.3 & 26.5  & 33.2  & 40.5  \\
    \midrule
    \textsc{DEGREE} & Generation
    & 55.4 & 62.1 & \textbf{65.8} & 65.8 & 68.3 & 68.2
    &  21.7 & 30.1 & 35.5 & 41.6 & 46.2 & 48.7 \\ 
    \textbf{ RAP$_{(DEGREE)}$ } & Generation
    & \textbf{59.3} &  \textbf{65.7}  & 65.2 & \textbf{67.1} & \textbf{70.0} & \textbf{69.7}
    & \textbf{23.5} & \textbf{31.5}  & \textbf{36.5} & \textbf{46.5} & \textbf{49.1}  & \textbf{49.8}   \\
\bottomrule
\end{tabular}
\label{table:low-resource-ace}
\end{table*}

\section{Experiments}
We conduct comprehensive experiments to evaluate the performance by answering the following research questions:
\begin{itemize}
    \item \textbf{RQ1}: How does our RAP plugged into previous approaches perform when competing with SOTA?
    \item \textbf{RQ2}: How do different key modules in our RAP framework contribute to the overall performance?
    \item \textbf{RQ3}: What are the benefits of RAP when integrating different types and amounts of knowledge?
    \item \textbf{RQ4}: How effective is the proposed RAP in extracting the different types of entities, relations, and events?
\end{itemize}

\subsection{Experiment Settings}
\subsubsection{\textbf{Dataset.}} 
As to the event extraction, we conduct experiments on the following popular benchmark:
\textbf{ACE05-E} with 599 English annotated documents. 
We use the same split and pre-processing step following the previous work~\cite{DBLP:conf/emnlp/WaddenWLH19,DBLP:conf/acl/LinJHW20}.
Apart from \textbf{ACE05-E}, we employ another event extraction dataset, \textbf{CASIE}~\cite{DBLP:conf/aaai/SatyapanichFF20} in the cybersecurity domain. 
For the relational triple extraction, we leverage two popular public datasets, \textbf{NYT}~\cite{DBLP:conf/pkdd/RiedelYM10}  and \textbf{WebNLG}~\cite{DBLP:conf/acl/GardentSNP17} to assess our method.

\subsubsection{\textbf{Evaluation Protocols.}} 
For constructing the low-resource setting in \textbf{ACE05-E}, we follow DEGREE~\cite{DBLP:conf/naacl/HsuHBMNCP22}, which generates different proportions (1\%,  3\%, 5\%, 10\%, 20\%, 30\%) of training data and uses the original development set and test set for evaluation. 
As for \textbf{CASIE}, we adhere to the preprocessing of earlier work~\cite{DBLP:conf/acl/0001LDXLHSW22}, and then randomly split the training data into 1\% and 10\%.
As regards relational triple extraction, we also generate the training data randomly, dividing it into 1\%, 5\%, and 10\%.

\begin{table*}[t!]

\caption{Model performance of Relational Triple Extraction models in the low-resource setting. We report the mean performance of micro $F_1$ scores (\%) over 5 different splits. The best numbers are highlighted in each column.}
\begin{tabular}{l|c|ccc|ccc}
    \toprule
    \multirow{2}{*}{\textbf{Model}} & \multirow{2}{*}{\textbf{Type}} & \multicolumn{3}{c|}{\textbf{WebNLG}} & \multicolumn{3}{c}{\textbf{NYT}} \\
    \cline{3-8}
    & & 1\% & 5\% & 10\% & 1\% & 5\% & 10\% \\
    \midrule
    \textsc{TPlinker} & Classification
    & 0.00 & 0.00 &  0.00
    & 6.29 & 76.67 & 80.11 \\
   \midrule
    \textsc{$RelationPrompt$} & Generation
     & 23.77  & 45.45  &  56.53
   &  54.37  &  63.80  & 66.58 \\
   \textbf{RAP$_{(RelationPrompt)}$} & Generation
     & \textbf{27.72} & \textbf{47.04} & 57.38
    & 57.19  & 66.79  & 69.39 \\
    \midrule
     \textsc{$PRGC$} & Classification
    & 0.00  & 40.79  & 57.36
    & 59.91  & 75.36  & 79.96 \\
    \textbf{RAP$_{(PRGC)}$ }  & Classification
   &12.69  & 45.10  &  \textbf{59.20}
    & \textbf{61.01}  & \textbf{78.17}  & \textbf{81.99} \\ 
\bottomrule
\end{tabular}
\label{table:low-resource-re}
\end{table*}

For event extraction, we use the same evaluation criteria in previous work \cite{DBLP:conf/emnlp/WaddenWLH19,DBLP:conf/acl/0001LXHTL0LC20,DBLP:conf/naacl/HsuHBMNCP22,DBLP:conf/acl/LinJHW20} and report the F1 score of trigger classification (\textbf{Trg-C}) and argument classification (\textbf{Arg-C}). 
\textbf{Trg-C} evaluates whether a trigger's offset and event type match the gold one, and \textbf{Arg-C} evaluates whether an argument's offset, event type, and role label all match the gold ones.
For the relational triple extraction, we follow \cite{DBLP:conf/acl/ZhengWCYZZZQMZ20} and an extracted relational triple is only regarded as correct if it is an exact match with ground truth.

\subsubsection{\textbf{Baselines for Comparison.}}
Since \ours is a pluggable approach that can be adapted to different methods, we select strong baselines and empower them with \ours.
\begin{itemize}
  \item \textsc{TANL} \cite{DBLP:conf/iclr/PaoliniAKMAASXS21}: a method converts event extraction as translation tasks between augmented natural languages.
  \item \textsc{Text2Event} \cite{DBLP:conf/acl/0001LXHTL0LC20}: a sequence-to-structure generation method that converts the input passage to a tree-like event structure.
  \item \textsc{DEGREE} \cite{DBLP:conf/naacl/HsuHBMNCP22}: an end-to-end method creates templates for each event type and builds event-specific prompts for targeted information generation.
  \item \textsc{PRGC} \cite{DBLP:conf/acl/ZhengWCYZZZQMZ20}: an end-to-end classification based model that utilizes global correspondence to tackle the Relation Triple Extraction task. 
  \item \textsc{RelationPrompt} \cite{DBLP:conf/acl/ChiaBPS22}: an end-to-end generation-based model for zero-shot relational triple extraction. In our paper, we omit the process to generate samples and use the relation extractor as the base model.
\end{itemize}
Apart from these models, we compare \ours with other popular methods, including  \textsc{OneIE} \cite{DBLP:conf/acl/LinJHW20}, \textsc{BERT\_QA} \cite{DBLP:conf/emnlp/DuC20} and \textsc{TPlinker} \cite{DBLP:conf/coling/WangYZLZS20}. 

\begin{table}[t!]
\caption{Low resource results for the cybersecurity dataset CASIE. The highest scores are in \textbf{bold}.}
\scalebox{0.93}{
\begin{tabular}{l|c|cc|cc}
    \toprule
    \multirow{2}{*}{\textbf{Model}} & \multirow{2}{*}{\textbf{Type}} & \multicolumn{2}{c|}{\textbf{1\% data}} & \multicolumn{2}{c}{\textbf{10\% data}}  \\
    \cline{3-6}
    & &  Tri-C &  Arg-C & Tri-C &  Arg-C  \\
    \midrule
    OneIE & Cls &  8.2  & 1.1  & 46.5 & 35.6 \\
    \midrule
    TANL & Gen &  3.8  & 10.1  & 50.3 &	37.3  \\
    \textbf{ \textsc{ RAP$_{(TANL)}$ } } & Gen & 1.7 & 14.4 & \textbf{53.6} & 37.4  \\
    \midrule
    \textsc{Text2Event} & Gen & 10.6  & 11.8 & 39.7 & 35.3 \\
    
    \textbf{ \textsc{ RAP$_{(TEXT2EVENT)}$ } } & Gen &  \textbf{12.0}  & \textbf{15.6}  & 47.6 & \textbf{39.1}  \\
    \bottomrule
\end{tabular}}

\label{table:low-resource-casie}
\end{table}


\subsection{Performance Comparison with SOTA (RQ1)}
\textbf{Low-resource}. We list the results of Event Extraction in Table~\ref{table:low-resource-ace} (ACE05-E) and Table~\ref{table:low-resource-casie} (CASIE), while the results of Relation Triple Extraction in Table~\ref{table:low-resource-re}. 
We can observe that \ours~ demonstrates strong competitiveness on both trigger classification and argument classification tasks. 
For the trigger classification task, \textsc{ RAP$_{(TEXT2EVENT)}$ } shows improvements in almost all settings compared with the base method \textsc{Text2Event}, while \textsc{RAP$_{(DEGREE)}$} outperforms all the other models except the 5\% setting in ACE05-E.
For the argument classification task, \ours~  surpasses all baselines in all the settings for ACE05-E. 
\ours~ also shows improvement in the cybersecurity domain. Compared with the base models \textsc{TANL} and \textsc{Text2Event} \cite{DBLP:conf/acl/0001LXHTL0LC20}, \ours~ achieves significant improvement in almost all settings except 1\% settings in \textsc{TANL}.

\begin{table}[ht]
\centering
\caption{Results on ACE05-E for event extraction in the supervised learning setting.}
\begin{tabular}{l|ccc|ccc}
    \toprule
    \multicolumn{1}{l|}{\multirow{2}{*}{\textbf{Model}}}  & \multicolumn{3}{c|}{\textbf{Trg-C}} & \multicolumn{3}{c}{\textbf{Arg-C}} \\  \cline{2-7}
    \multicolumn{1}{c|}{} & P & R & F1 & P & R & F1 \\
    \midrule
    \textsc{TANL}& - & - & 68.5 & - & - & 48.5 \\
    \textsc{Text2Event} & \textbf{69.6} & 74.4  & 71.9 & 52.5 & 55.2 & 53.8  \\
    \textsc{BART-GEN} & 69.5 & 72.8 & 71.1 & \textbf{56.0} & 51.6 & 53.7 \\
    \textsc{DEGREE-e2e} & - & - & \textbf{73.3} & - & - & 55.8  \\
    \textbf{\textsc{RAP$_{(DEGREE)}$}} & 66.5  & \textbf{79.6} & 72.5 & 53.5 & \textbf{58.7} & \textbf{56.0}  \\
    \bottomrule
\end{tabular}

\label{tab:supervised}
\end{table}%

\begin{table}[ht]
\caption{Results on NYT and WebNLG for relation triple extraction in the supervised learning setting.}
\begin{tabular}{l|ccc|ccc}
    \toprule
    \multirow{2}{*}{\textbf{Model}}
      & \multicolumn{3}{c|}{\textbf{WebNLG}} & \multicolumn{3}{c}{\textbf{NYT}} \\ \cline{2-7}
     & P & R & F1 & P & R & F1  \\
    \midrule
    \textsc{NovelTagging}   & 52.5 &  19.3 &  28.3 & 32.8 & 30.6 &  31.7 \\
    \textsc{MultiHead} & 57.5 & 54.1 & 55.7 & 60.7 & 58.6 & 59.6 \\
    \textsc{ETL-span}& 84.3 & 82.0 & 83.1 & 85.5 & 71.7 & 78.0  \\
    \textsc{RSAN} & 80.5 & 83.8 & 82.1 &  85.7 & 83.6 & 84.6 \\
    \textsc{TPLinker} & 88.9 & 84.5 & 86.7 & 91.4 & \textbf{92.6} & 92.0 \\
    \textsc{PRGC} &89.9 & \textbf{87.2} & 88.5 & \textbf{93.5} & 91.9 & \textbf{92.7} \\
    \textbf{\textsc{RAP$_{(PRGC)}$}} & \textbf{90.4} & 87.1 & \textbf{88.7}& 93.1 & 91.1 & 92.1 \\
    \bottomrule
\end{tabular}

\label{tab:supervised-re}
\end{table}%

As regards the relational triple extraction task, we evaluate \ours~ on both the generation-based model and the classification-based model, and from the table, we can observe a significant increase in both the WebNLG and NYT datasets. 
\textsc{ RAP$_{(RelationPrompt)}$} averages a 3.75\% improvement in all settings of the two datasets, and \textsc{ RAP$_{(PRGC)}$} outperforms all the other methods.  When applying classification-based models to low-resource scenarios, they can perform extremely badly. 
For instance, PRGC performs 0.00 in 1\% settings of WebNLG, while the performance rises up to 12.69\% with \ours~, and it shows the same tendency in the NYT dataset, which proves the effectiveness of our design.

\textbf{Fully-supervised}. We also report the performance in the high-resource setting for controlled comparisons. 
Table~\ref{tab:supervised}  shows the results of high-resource event extraction and Table~\ref{tab:supervised-re} shows the results of high-resource relation triple extraction.
For event extraction tasks, our method, \textsc{RAP}, achieves slightly better performance than previous methods on argument extraction (ACE05-E Arg-C) and relational triple extraction (WebNLG)
However, the advantage of \textsc{RAP} becomes less obvious for event trigger detection (ACE05-E Tri-C) and relational triple extraction (NYT) when sufficient training examples are available.

\begin{table}[ht]
\centering
\small
\renewcommand{\arraystretch}{1.2} 
\caption{
Results (F1-score, \%) of EE on ACE05-E with different knowledge types. 
We select the settings of 1\% and 3\% data. The backbone model is DEGREE.}
\begin{tabular}{l|c|c|c|c}
\toprule 
\multirow{2}{*}{\textbf{Method}}
    & \multicolumn{2}{c|}{\textbf{1\% Data}}  & \multicolumn{2}{c}{\textbf{3\% Data}}   \\  \cline{2-5}
    & Tri-C & Arg-C & Tri-C & Arg-C  \\
\midrule

\textbf{\textsc{RAP}} & \textbf{59.3} & \textbf{23.5} & \textbf{65.7 }& \textbf{31.5} \\
\midrule
    w/o Instances     & 58.1 & 22.4 & 63.6  & 31.1 \\
    w/o Trigger Info. & 53.1 & 15.2 & 60.9 & 28.7 \\ 
    w/o Argument Info.      & 53.1 & 9.9 & 60.6 & 23.7 \\
    w/o Type Struct.    &  57.5 & 20.2 & 62.0 & 30.4 \\
\bottomrule
\end{tabular}

\label{tab:ablation-event}
\end{table}

\begin{table}[ht]
\centering
\renewcommand{\arraystretch}{1.2} 
\caption{
Results (F1-score, \%) of relational triple extraction with different knowledge types. 
We select the settings of 1\% and 5\% data. The backbone model is PRGC.}
\begin{tabular}{l|c|c|c|c}
\toprule 
\multirow{2}{*}{\textbf{Method}}
    & \multicolumn{2}{c|}{\textbf{1\% Data}}  & \multicolumn{2}{c}{\textbf{5\% Data}}   \\  \cline{2-5}
    & NYT & WebNLG & NYT & WebNLG  \\
\midrule

\textbf{\textsc{RAP}} & \textbf{60.75} & \textbf{13.59} & \textbf{78.74}& \textbf{47.26} \\
\midrule
    w/o Instances     & 59.89 & 8.77 & 77.07 & 46.75 \\
    w/o Relation Info.  & 59.48 & 10.69 & 77.79 & 46.05 \\
    w/o Structure Info.    & 59.95 & 9.87 & 75.52 & 44.85 \\
\bottomrule
\end{tabular}
\label{tab:ablation-relation}
\end{table}

\subsection{Ablation Study of RAP Framework (RQ2)}

In this part, we present extensive ablation studies to support our design.
To better understand the contribution of each component in the prompt, we ablate \ours~ for both relational triple extraction and event extraction tasks.
Table~\ref{tab:ablation-event} lists the results of ACE05-E and Table~\ref{tab:ablation-relation} illustrates the results of WebNLG and NYT. 
We discover that nearly all forms of information are essential since their absence has a detrimental effect on performance.
For all tasks, we notice a reduction in performance when text instances are omitted from the prompts. 
For the event extraction task, among different components of prompts, the argument information has a great impact on the performance of both Tri-C and Arg-C. Removing the argument information from the prompt leads to a huge performance drop. 
With regard to the relation triple extraction task, the removal
of relation information and structure information leads to performance drops, which also validates their necessity.
What's more, when less training data is provided, the advantage of any of these components becomes more apparent.

\begin{figure*}
    \centering
    \includegraphics[width=0.93 \textwidth]{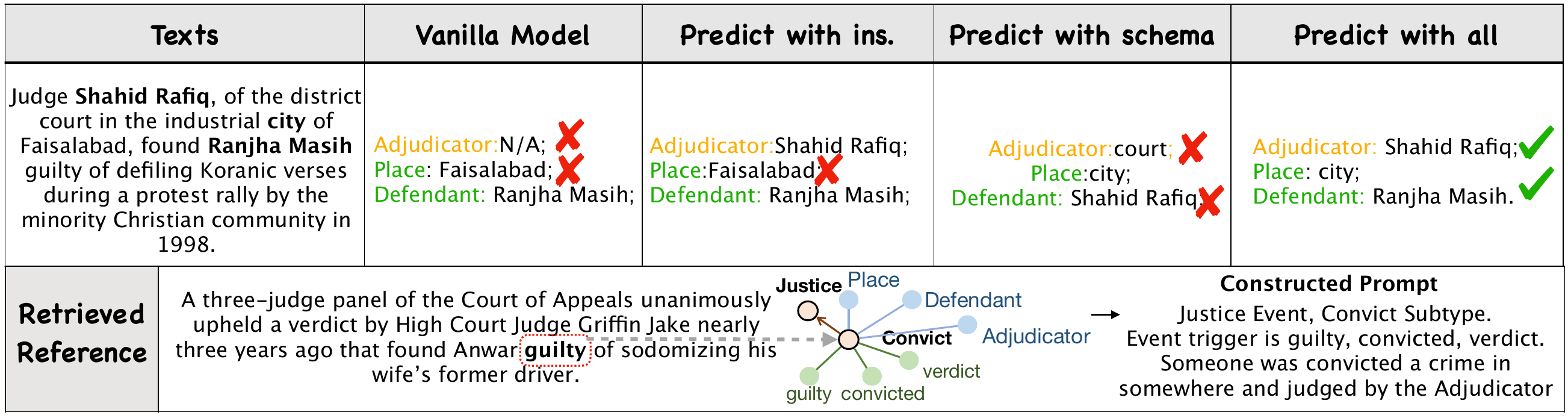}
   \caption{
   A case study in ACE05-E to analyze the effect of text instances $\mathcal{I}$ and schema graph $\mathcal{G}$ on the Argument Classification task. 
   Vanilla model didn't utilize the reference as prompt. Predict with ins. only use the text instance and Predict with schema only use the schema graph in the retrieved reference.}
   \label{table:case}
\end{figure*}

\subsection{Benefits of RAP with different type and amount of knowledge (RQ3)}

\begin{figure}
    \flushleft
    \includegraphics[width=0.48\textwidth]{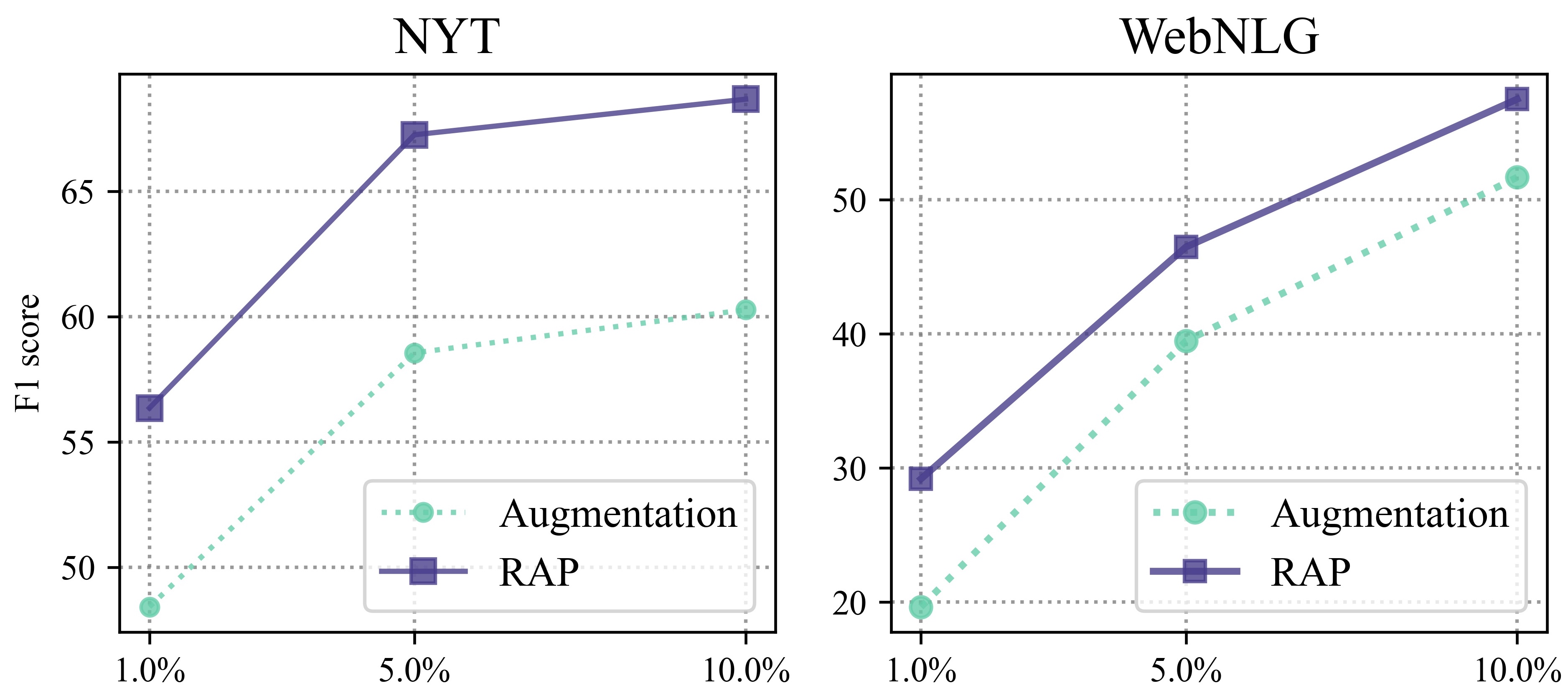}
    \caption{
    Comparison with Data-augmentation method.
    The x-axis is the percent of the training data.
    The backbone model is RelationPrompt.}
    \label{fig:data-augment}
\end{figure}
\textbf{Relevant data \& schema as references (prompts) outperforms data augmentation with retrieved instances}.
To determine whether the improvements can indeed be attributed to the architecture of the reference store or simply the additional data (weak supervised data), we compare our model RAP to the data augmentation method.
In detail, getting the retrieved entries $d = (c,\tilde{y},p)$, we transform them into the same format as training data. 
The query is $c$, and the label is paraphrased from the schema subgraph that is pointed to $p$.
Then, we train our model with both the training data and the retrieved references. 
We conduct experiments on two triple extraction datasets and show the results in Figure~\ref{fig:data-augment}. 
We can find that \ours~ outperforms the data augmentation method under both datasets, which verifies the effectiveness of the prompt. 
One possible reason may be that our model can dynamically select relevant knowledge (instances) as an external prompt, which will not change the original semantics of the input sequence. 
However, using those retrieved instances as data augmentation may introduce noise for training, thus, leading to performance decay.

\textbf{Similar examples contribute to the context understanding and schemas play a more essential role}.
To further understand the interaction between these two types of knowledge, we conducted a case study to investigate how text messages and schema information complement each other and the specific information provided by each type of reference. 
We select an instance from the ACE05-E task.
As shown in Figure~\ref{table:case}, the sentence here describes a \emph{Convict} event and contains complicated information.
The argument and role contain the \emph{Adjudicator}, \emph{Place}, and the \emph{Defendant}. 
From the figure, we can find the vanilla model failed to retrieve the \emph{Adjudicator} and \emph{Place}.
\begin{figure}
    \flushleft
    \includegraphics[width=0.5\textwidth]{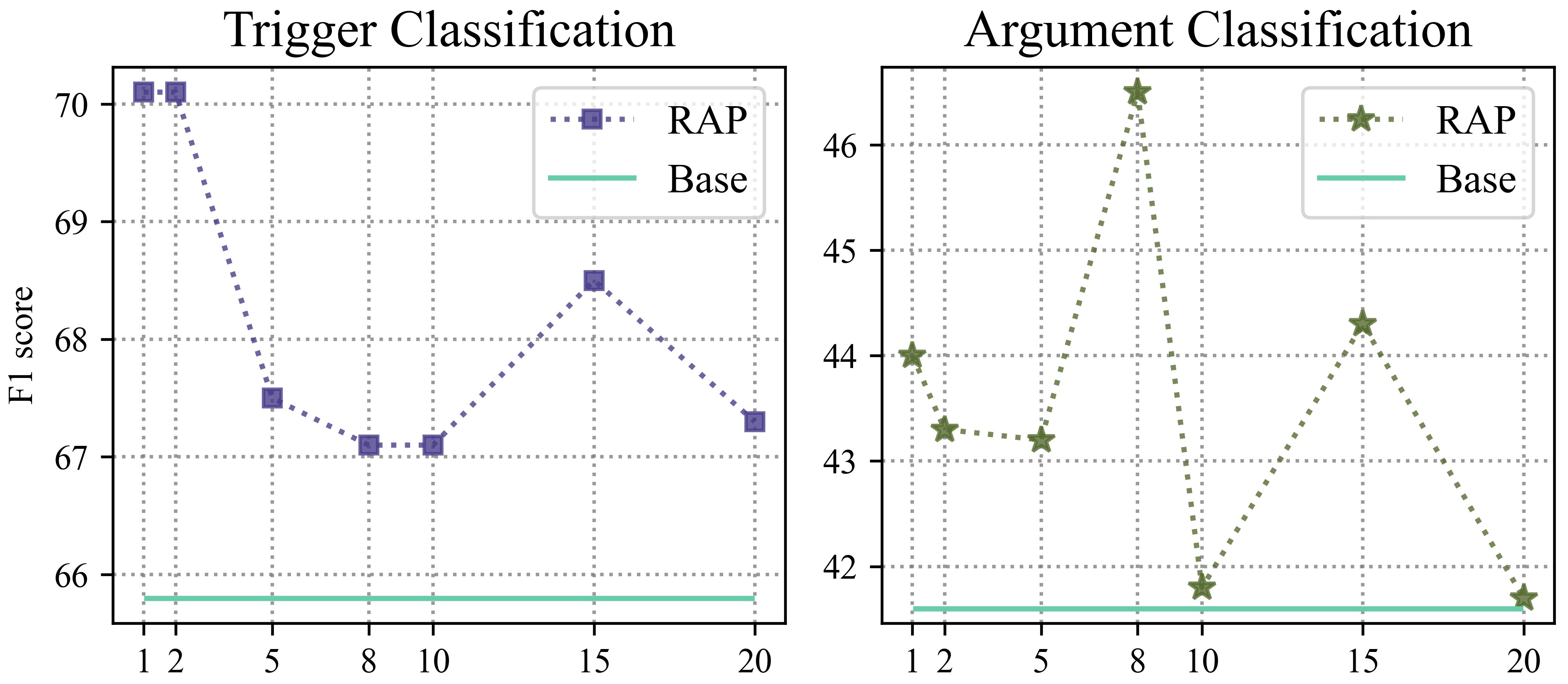}
    \caption{
    Effect of the number of instances retrieved on the model's performance for the ACE05-E task under 10\% setting. 
    The base model means we do not utilize the reference as the prompt. 
    The x-axis is the number of instances.}
    \label{fig:number}
\end{figure}
\begin{figure*}
    \centering{
    \includegraphics[width=0.97 \textwidth]{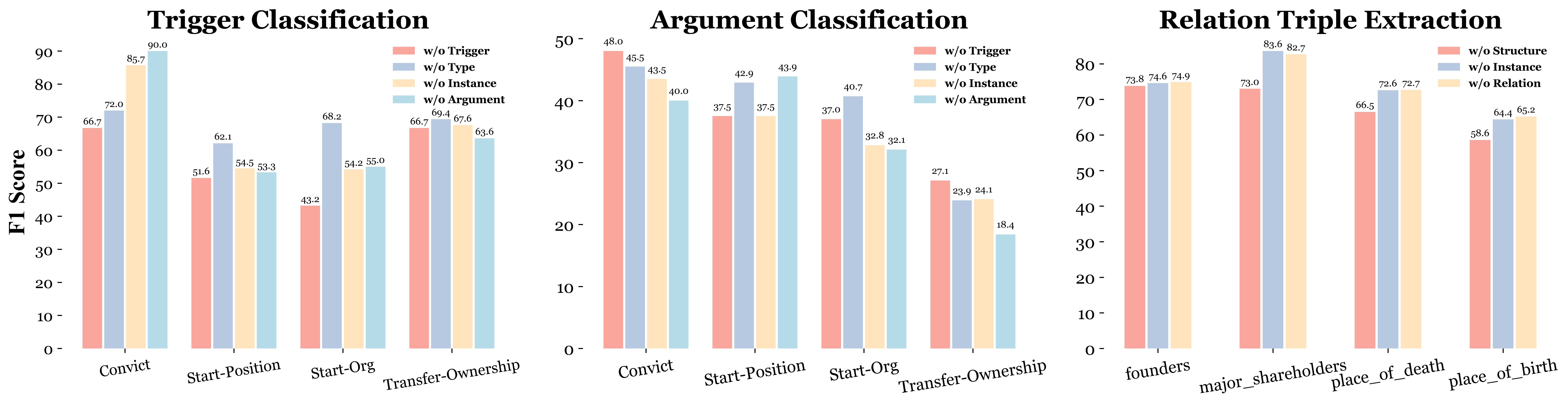}
    }
    \caption{F1-score of four event types based on various knowledge inputs on ACE05-E under 10\% setting.}
    \label{fig:event_type}
\end{figure*}

One interesting phenomenon is that similar texts improve the model's understanding of the downstream tasks.
When the model simply adopts the text instance in the retrieved reference, it correctly identifies the \emph{Adjudicator}.
Moreover, the text instance can assist the model in better understanding the schema.
If we simply utilize the schema information, the model incorrectly infers that \emph{`court'} is the \emph{Adjudicator}. 
However, after injecting the text instance, the model makes a correct prediction for all the arguments.
To be specific, the retrieved sentence also includes the event of \emph{Convict}. 
Despite the retrieved sentences having different arguments and roles, it shares the same structure with our input \nw{judge \blank\xspace found \blank\xspace the guilty of \blank\xspace}, which implicitly indicates the \emph{Convict} event structure and schema. 
These similar examples enhance the model's understanding of the event using only a few samples, due to our method's ability to better capture dependencies.

\textbf{Using more retrieved knowledge data can only boost performance to a certain extent, not continuously better due to negative knowledge fusion.}
We further conduct experiments to analyze how the number of retrieved references ($K$) affects performance. 
We take the ACE-05E task as an example.
As shown in Figure~\ref{fig:number}, the model performs best when we utilize the top 1–2 chosen references for Tri-C task.
The model benefits from knowledge but faces noise with more retrieved references.
Lower similarity in later references may cause noise, affecting performance. 
Arg-C follows a similar trend, peaking at around 8 references, as argument classification is more challenging and needs more similar references for learning.

\subsection{Different Type Analysis of Entity, Relation and Event (RQ4)}
The above-mentioned experiments prove the effectiveness of our method while the utility of the prompt may vary in different cases. 
To better understand the principle of knowledge injection under the low-resource scenario, we analyze the effects of the prompts on different event types and relation types.

For the event extraction, we select four event-types that appear less than five times, namely ``\textit{Start-Position}'', ``\textit{Convict}'', ``\textit{Transfer-Ownership}'' and ``\textit{Start-Org}''. For the relational triple extraction, we also select four types including: ``\textit{founders}'', ``\textit{major\_sharehoders}'', ``\textit{place\_of\_death}'', and ``\textit{place\_of\_birth}''.
Figure~\ref{fig:event_type} demonstrate the F1 score of all these target types based on various forms of prompt input.
We observe that:
(1) For the event extraction task, different components of the \textsc{RAP} show different effects on both tasks. Overall, trigger information plays a more vital role in the trigger classification task, while the instance and arguments are more significant for the argument classification task. 
(2) Event type has less influence on the Trig-C for the ``\textit{Start-Position}'' and ``\textit{Start-Org}'' event type, probably because these event type is less inductive and contains little information of the event triggers. 
(3) The performance of Arg-C on ``\textit{Convict}'', ``\textit{Transfer-Ownership}'' and ``\textit{Start-Org}'' types is greatly affected by the arguments and instances. 
(4) Unlike Event Extraction, different parts of the prompt demonstrate similar trends on these different types of Relation Triple Extraction: triple structure is the most important part of the prompt, while instance and relation information are not that influential.

\section{Related work}
\textbf{Relational Triple Extraction.}
Early works \cite{DBLP:conf/acl/ChanR11} apply the pipelined methods to perform relation classification after extracting all the entities.
\cite{DBLP:conf/coling/WangYZLZS20} employs a token pair linking scheme which performs two matrix operations for extracting entities and aligning subjects with objects under each relation of a sentence.
The recent well-performed model PRGC \cite{DBLP:conf/acl/ZhengWCYZZZQMZ20} is an end-to-end classification model that leverages a global correspondence matrix. 
Generation-based models \cite{DBLP:conf/acl/ChiaBPS22} also emerged with strong performance. 
However, few works consider the prompt to enhance the model for this complicated task. 
In this work, we utilize schema-aware references as prompts \textsc{RAP} to enhance the relation triple extraction task.

\textbf{Event Extraction.}
Early studies formulate Event Extraction as token-level classification,to identify triggers and arguments in texts.
Numerous studies~\cite{DBLP:conf/naacl/NguyenCG16, DBLP:conf/acl/YangFQKL19, DBLP:conf/emnlp/WaddenWLH19,DBLP:conf/acl/LiJH13,DBLP:conf/naacl/YangM16} employ pipeline-style frameworks for this task.
Meanwhile, some work casts event extraction as a machine reading comprehension (MRC) problem~\cite{DBLP:conf/emnlp/LiuCLBL20,DBLP:conf/emnlp/DuC20,DBLP:conf/emnlp/LiPCWPLZ20}. 
They construct question-answer pairs to query event triggers and arguments.
Recently, many generation-based models have been proposed~\cite{DBLP:conf/acl/0001LXHTL0LC20,DBLP:conf/iclr/PaoliniAKMAASXS21, DBLP:conf/emnlp/HuangTP21, DBLP:conf/acl/HuangHNCP22, DBLP:conf/naacl/LiJH21,DBLP:conf/naacl/HsuHBMNCP22,DBLP:conf/icassp/SiPLXL22}.
The generation-based model is more flexible and portable, reducing the burden of annotation and can extract triggers and arguments simultaneously.

\textbf{Retrieval Augmented Models.}
Retrieval-augmented models have been applied to Language Model (LM) ~\cite{DBLP:conf/iclr/KhandelwalLJZL20}, text generation~\cite{DBLP:journals/corr/abs-2202-01110,DBLP:journals/corr/abs-2010-04389} and open-domain question answering~\cite{DBLP:conf/nips/LewisPPPKGKLYR020,DBLP:journals/corr/abs-2002-08909}. 
More works adopt retrieval-augmented model to tackle other tasks such as question answering~\cite{DBLP:journals/corr/abs-2204-04581}, knowledge graph completion~\cite{DBLP:journals/corr/abs-2201-05575}, relation extraction~\cite{DBLP:conf/sigir/ChenLZTHSC22} and  NER~\cite{DBLP:journals/corr/abs-2203-17103}. 
\citet{DBLP:conf/icml/0002XHSRN22} propose RETOMATON via a neuro-symbolic synergy of neural models with symbolic automata.  
Recently, \citet{DBLP:conf/acl/WangXFLSX0022} noticed that retrieving examples from training data can enhance the model performance for different NLU tasks.
However, few works apply retrieval methods for event extraction and relation triple extraction tasks.
Unlike those approaches, we focus on knowledge graph construction and propose \textsc{RAP} with a schema-aware reference store and conduct retrieval method to enhance the model.

\section{Conclusion and Future Work}
In this paper,  we propose \textsc{RAP} for data-efficient knowledge graph construction, which constructs a schema-aware reference store and dynamically selects informative knowledge as prompts for integration. 
Experimental results demonstrate that our model achieves competitive results with current-state models for both event extraction and relation triple extraction tasks.
\textsc{RAP} can be applied to different existing methods. 
Additionally, we provide an in-depth analysis when injected with different components of the prompt.
In the future, we plan to 1) explore more symbolic knowledge, such as axiom rules for knowledge graph construction, 2) extend our approach to general natural language generation tasks. 
\begin{acks}
We would like to express gratitude to the anonymous reviewers for their kind comments. 
This work was supported by the National Natural Science Foundation of China (No.62206246 and U19B2027), Zhejiang Provincial Natural Science Foundation of China (No. LGG22F030011), Ningbo Natural Science Foundation (2021J190), and Yongjiang Talent Introduction Programme (2021A-156-G), CAAI-Huawei MindSpore Open Fund, and NUS-NCS Joint Laboratory (A-0008542-00-00).
This work was supported by Information Technology Center and State Key Lab of CAD\&CG, ZheJiang University.
\end{acks}



\bibliographystyle{ACM-Reference-Format}
\bibliography{anthology, custom}

\end{document}